# Persian Rhetorical Structure Theory

Sara Shahmohammadi
MSc Graduate
Faculty of New Sciences and Technologies            sshahmohamadi@ut.ac.ir
University of Tehran

Hadi Veisi*
Assistant Professor                                 h.veisi@ut.ac.ir
Faculty of New Sciences and Technologies
University of Tehran

Ali Darzi                                           alidarzi@ut.ac.ir
Professor
Faculty of Literature and Humanities University
of Tehran

**Abstract**

Over the past years, interest in discourse analysis and discourse parsing has steadily grown, and many discourse-annotated corpora and, as a result, discourse parsers have been built. In this paper, we present a discourse-annotated corpus for the Persian language built in the framework of Rhetorical Structure Theory as well as a discourse parser built upon the DPLP parser, an open-source discourse parser. Our corpus consists of 150 journalistic texts, each text having an average of around 400 words. Corpus texts were annotated using 18 discourse relations and based on the annotation guideline of the English *RST Discourse Treebank* corpus. Our text-level discourse parser is trained using gold segmentation and is built upon the DPLP discourse parser, which uses a large-margin transition-based approach to solve the problem of discourse parsing. The performance of our discourse parser in span (S), nuclearity (N) and relation (R) detection is around 78%, 64%, 44% respectively, in terms of F1 measure.

## 1. Introduction

Discourse can be simply defined as the patterns utterances follow and discourse analysis as the study of these patterns (Jorgensen and Phillips, 2002). It is by means of these patterns that coherence forms in a text, and we manage to comprehend a text beyond a set of individual sentences. Rhetorical Structure Theory (Mann and Thompson, 1987) or RST is one of the existing approaches aiming at studying these patterns. It tries to formalize and explain how coherence develops in a text. To do so, it characterizes text "structure through relations that hold between parts of the text." (Mann and Thompson, 1988, p. 243).

RST discourse parsing refers to the extraction of RST structure from a text. Even though discourse parsing is considered to be a challenging task, it is a prospective area, since it can be beneficial to a range of other tasks; such as text summarization, essay grading and argumentation mining (Taboada and Mann, 2006).

RST-annotated corpora have been built for various languages so far; for English, for in-stance, there are *RST Discourse Treebank* (Carlson et al., 2003) and *GUM* corpus (Zeldes, 2017). There are also RST-annotated corpora for Spanish (Da Cunha et al., 2011), Russian (Toldova et al., 2017), Dutch (Van Der Vliet et al., 2011), German (Stede, 2004), Basque (Iruskieta et al., 2013), and Bangla (Das and Stede, 2018). RST-annotated corpora could be of great use in developing natural language processing tools; especially when struggling with beyond sentence-level structures. That is why we have taken the first step and created an RST-annotated corpus for the Persian language. Although there is already a discourse cor- pus for Persian, known as *Persian Discourse Treebank and Coreference Corpus* (Mirzaei and Safari, 2018) and annotated in the framework of Penn Discourse Treebank, it only marked sentence-level relations and was unfortunately not yet published and publicly available as we built Persian RST corpus, otherwise it would have been possible to select Persian RST corpus texts differently. Persian RST corpus currently consists of 150 journalistic texts, annotated using 18 discourse relations and covering a variety of discourse structures. In addition, we have also built a discourse segmenter and parser based on the open-source Discourse Parsing from Linear Projection or (DPLP) (Ji and Eisenstein, 2014). This paper is a report of what has been done so far to build this corpus and this parser.

In Section 2, we will have a brief look at fundamentals of RST, RST-annotated corpora, and some discourse parsers built so far. In Section 3, we will discuss our Persian RST corpus and the annotation process. In Section 4, we will introduce our discourse parser trained with the help of the open-source DPLP parser. Lastly, in Section 5, we will present the summary and the conclusion.

## 2. Background and Related Works
### 2.1. Rhetorical Structure Theory

Rhetorical Structure Theory emerged out of a natural language generation project at In- formation Sciences Institute in the late 1980s (Mann and Thompson, 1987, p. 2). It mainly concerns itself with formalizing and explaining the discourse structure in a text. In other words, it tries to clarify and model how we understand and make sense of texts beyond sentence-level. Of course, RST could also be used to study discourse structure in other forms of language like dialogues, however, the focus has been primarily more or less on texts so far.

There are some key concepts in RST, which help formalize and model discourse struc- ture; including *discourse units*, *discourse relations*, and *nuclearity*. We will now have a brief look at each of these key concepts.

In RST, discourse structure is explained using discourse relations holding and gluing dis- course units together. Discourse units could be either *Elementary Discourse Units* (*EDUs*) or *discourse spans*. According to Stede (2011, p. 87.), quoting Polanyi (2004), EDUs are "the syntactic constructions that encode a minimum unit of meaning and/or discourse function interpretable relative to a set of contexts". Discourse units join each other in a recursive process to consequently form larger and more complex units, namely discourse spans, until we reach a span that covers and "contains a set of text spans that constitute the entire text" (Mann and Thompson, 1987), namely the *root span*.

*Discourse relations* describe how discourse units are related in terms of discourse; e.g. discourse units can be in a sort of contrast, cause or a temporal relation to each other. There is obviously no exhaustive set of discourse relations yet. Various relation sets have been proposed so far and it's always possible to modify these relation sets when needed.

Another key concept in RST is the concept of nuclearity. When two discourse units join each other to form a new unit, one could play a more major role and be more essential than the other; that is, if this unit is deleted, the text will become incoherent. In this case, the more essential unit would be the nucleus, the other unit would be the satellite, and the relation would be a mononuclear discourse relation. It is also possible that both units are of equal importance. In this case, they are both marked as nucleus, and the relation would be a multinuclear discourse relation. There are some ways to distinguish the nucleus from the satellite; e.g. Carlson and Marcu (2001, p. 34) propose deletion test and replacement test to do so. While these tests are of great help, there are still cases where it is not possible to decide on nuclearity by relying on them. This is probably one of the challenges in Rhetorical Structure Theory, and why various texts could result in different RST-trees.

Figure 1 shows a simple RST tree of Example 1. In this example, A is the nucleus and B the satellite. The two EDUs join each other by an *Enablement* relation and form a discourse span.

**Example 1**

[باید لایحه اعطای تابعیت به فرزندان مادران ایرانی تصویب شود]A[تا تابعیت از مادر به فرزند برسد]B (Zandrazavi, 2019).

[The bill to give citizenship to children of Iranian mothers must get approved]A[in order for citizenship to pass on from mother to child]B

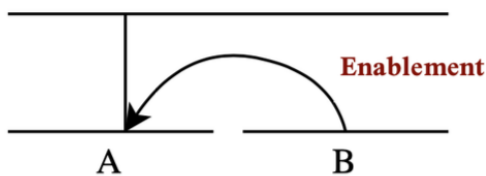

Figure 1: RST Tree of Example 1

## 2.2. Rhetorical Structure Theory Corpora

Discourse-annotated corpora in the framework of Rhetorical Structure theory have been built for a variety of languages; such as English (Carlson et al., 2003), German (Stede, 2004), Dutch (Van Der Vliet et al., 2011), Spanish (Da Cunha et al., 2011), Russian (Toldova et al., 2017), Portugese (Cardoso et al., 2011), Chinese (Ming, 2008) and recently for Bangla (Das and Stede, 2018) and Basque (Iruskieta et al., 2013). These corpora all follow RST principles (such as the concept of EDUs and merging EDUs to form spans, the concept of discourse relations, or the assumption that a tree structure can be used to model discourse structure), however, there are some variations in the annotations; e.g. while the English *RST Discourse Treebank* (Carlson et al., 2003) has a list of relations with 78 fine-grained relations, categorised in 16 coarse-grained relations, Spanish RST corpus has a shorter list of relations; texts in German RST corpus (*PCC*) are annotated with six layers of information, such as syntax and co-reference information (Stede, 2004, pp. 96-97), and the Dutch RST corpus has also been annotated with lexical cohesion information (Van Der Vliet et al., 2011).

## 2.3. Text-level Discourse Parsing

Discourse segmentation can be formulated either as a -mostly binary- classification problem and solved using classification algorithms; such as *Support Vector Machine* (*SVM*) or Logistic Regression, or as a sequential labeling problem and solved by algorithms such as *Conditional Random Field* (*CRF*) (Feng and Hirst, 2014b, p. 512-513). RST discourse parsing is building an RST-tree from a set of EDUs. It involves merging discourse units by means of proper discourse relations, until we reach the root node, and it consists of a number of decisions:

- EDU formation (which spans/EDUs should join each other?)
- Nuclearity detection (which span/EDU is the nucleus and which is the satellite?)
- Relation detection (which coherence relation holds between the units?).

It is possible to train separate classifiers for each of these tasks, however, it is also possible to train a classifier that decides for span formation and nuclearity at the same time, and train another classifier to detect the coherence relations. Moreover, a tree-building algorithm should also be chosen in order to build the discourse tree. Tree-building can be done through a greedy bottom-up approach or shift-reduce parsing. It is also possible to use *CKY*-like algorithms jointly with probabilistic models to solve the problem of discourse parsing; which is what Joty et al (2013) have done.

DPLP (Ji and Eisenstein, 2014) is an open-source discourse parser using an SVM model as well as a shift-reduce parser. It mainly focuses on modeling lexical features related to discourse parsing, since alternative lexicalizations are what make discourse parsing so complicated. As the first step for RST discourse parsing in Persian, we have used DPLP, and we have trained a parser using Persian RST corpus.

### 3. Persian Rhetorical Structure Theory Discourse Treebank

This corpus is the first RST-annotated corpus built for the Persian language. In this section, we are going to talk about the stages and the process of corpus building and annotation. First, we explain about text selection. We will then discuss discourse annotation; a brief account of how we segmented and labeled the texts as well as discussing some problematic cases.

Table 1: Corpus Characteristics

| | Documents | Average Words per Document | EDUs | Spans | Nuclei | Satellites |
|---|---|---|---|---|---|---|
| Etemad | 41 | 477 | 2667 | 2390 | 3512 | 1545 |
| Shargh | 60 | 561 | 1983 | 1788 | 2624 | 1147 |
| Meidan +1* | 49 | 241 | 1139 | 1000 | 1511 | 628 |
| Corpus | 150 | 409 | 5789 | 5178 | 7647 | 3320 |

\* The one single document taken from Iran Transportation Magazine is included here.

**3.1. Text Selection**

Although RST could be used to analyze a variety of text types, we decided to limit the corpus to editorials and short, analytic journalistic texts mostly categorized as social and political texts. This decision was mainly inspired by the other RST corpora, like the English *RST Discourse Treebank* corpus (Carlson and Marcu, 2001). We chose the texts from the following sources: Etemad newspaper[1] (41 texts), Shargh newspaper[2] (60 texts), Meidan website[3] (48 texts) and one text taken from Iran Transportation Magazine[4].

Table 2: Number of Texts from Each Source

|  | Corpus | Train Set | Development Set | Test Set |
|---|---|---|---|---|
| Etemad | 60 | 52 | 4 | 4 |
| Shargh | 41 | 35 | 3 | 3 |
| Meidan +1 | 49 | 43 | 3 | 3 |
|  | 150 | 130 | 10 | 10 |

Text selection from the sources was not done randomly since text quality and coherence could not be automatically checked while there were plenty of hasty or at times partially incoherent texts. Therefore, before including each text in the corpus, we had to ensure the text is of acceptable quality.

Finally, we collected 150 texts. Texts selected from Shargh and Etemad newspapers are mostly editorials and are longer compared to Meidan texts. The average number of words per text is 406 words; the shortest text having a length of 115 words and the longest having a length of nearly 800 words. Table 1 and Table 2 summarize our corpus characteristics.

### 3. 2. Annotation

We have followed Carlson and Marcu tagging manual (Carlson and Marcu, 2001). The tool used for annotation was *rstweb* (Zeldes, 2016). RST trees are saved in xml files. In order to reduce complexity, only one level of embedding has been marked. Corpus documents have been first normalized and then annotated. The normalization consisted of unifying different forms of some letters and words. Regarding character normalization, different forms of ی, were unified. For instance, different forms of the character ک and و, ی, characters like which are (ي/ی/ی/ی/ي), were all turned into ی. Regarding unifying word forms, you can find some normalization cases in Table 3. It should be noted that we have added spaces around punctuation marks to facilitate the tokenization and segmentation process.

---

[1] etemadnewspaper.ir
[2] sharghdaily.com
[3] meidaan.com
[4] iran-transportation.com

| Table 3: Text Normalization | |
|---|---|
| Normalized Form | Raw Form |
| کتاب‌ها | کتاب ها |
| توانایی‌ها | توانایی ها |
| می‌روم | می روم |
| می‌خورم | می خورم |
| است ، | است، |
| می‌دانی ؟ | می‌دانی؟ |

The annotation process consisted of four types of decisions: how to choose elementary discourse unit (EDU) boundaries or segmentation, how to form a span, how to assign nuclearity, and how to decide which relation holds between two spans. In what follows, we discuss each of these decisions and state our criteria in annotation.

### 3.2.1. Segmentation

Each text was first segmented into elementary discourse units. Various researchers have defined elementary discourse units differently; some take clauses as EDUs while others take sentences or prosodic units as EDUs (Carlson and Marcu, 2001). In our work, elementary discourse units are considered to be clauses. Phrases can also be marked as elementary discourse units only if they contain a strong discourse marker. Although the segmentation step is more or less clear, there are still tricky cases that can cause inconsistency throughout the corpus. The most important source of inconsistency lies in deciding about the valency of a verb and how many arguments a verb gets. What gets problematic here almost always concerns the question "Does this verb get a clause as an argument or not?"; if it does, we do not mark the clause as a separate segment and if it does not, we do. An example of such a case is a category of verbs that Tabibzadeh (2006) calls "impersonal verbs with a null subject". In Example 2, you can see an example of such verbs.

**Example 2**
لازم بودن
to be highly needed
[لازم است به این نکته توجه کنی]
[It is highly needed that you pay attention to this]

Another example is what Tabibzadeh (2006) calls "compound linking verbs". An example is shown in Example 3.

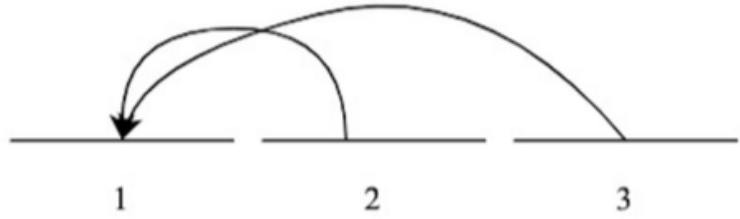

Figure 2: An Invalid Schema in Persian RST Treebank

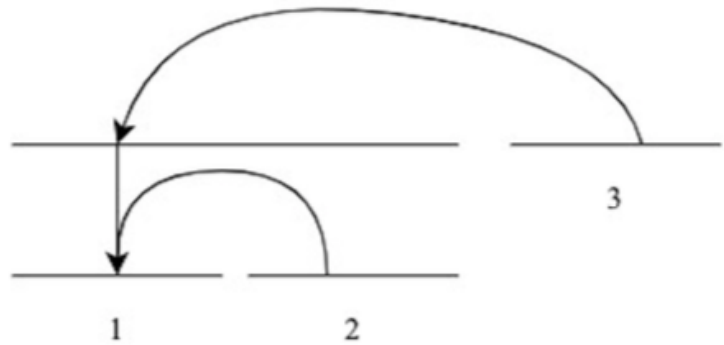

Figure 3: How the Graph in Figure 2 is Annotated in Persian RST Treebank

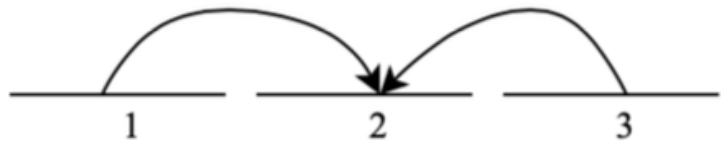

Figure 4: A Schema Not Used in Persian RST Treebank

**Example 3**
باعث شدن
to cause
[آن‌ها باعث شدند این آتش شعله‌ور شود]
[They caused the fire to flare up]

      To check for verb valency, we consulted Tabibzadeh (2006). Besides, we also used the syntactic valency lexicon for Persian (Rasooli et al., 2011). Of course, we still encountered verbs

that could not be found in either of these sources, so we had to resort to our intuition at times.

### 3. 2. 2. Forming Spans

In mononuclear relations, there are always only two discourse units that enter a relation to form a new span; that is, we do not allow the tree corresponding to the Figure 2. Instead, after the first relation, a span is created and that span enters a relation with the third unit. Like the graph shown in Figure 3.

Also, we do not allow an *SNS* (Satellite-Nucleus-Satellite) schema like the one shown in Figure 4; that is, we do not allow two satellites to simultaneously and at the same level get related to a nucleus. We either merge 1-2 or 2-3 first, and then add the other satellite on top of it.

### 3. 2. 3. Nuclearity Assignment

Nuclearity was also challenging to assign at times, especially in larger spans. This is not surprising since all judgements in RST annotation are plausibility judgements (Mann and Thompson, 1988), so various annotations are possible for a text. The rule of "nucleus can stand independently on its own, while satellite would lose meaning if nucleus is omitted" does not always work for nucleus detection. So, when assigning nuclearity in larger spans, we tried to disambiguate by asking the question "Which span would we want to include in the summary of this text?". By answering this question, we tried to remain as consistent as possible and pick the most salient span. Obviously, it is recommended that we formulate more objective and concrete criteria further on to avoid inconsistencies as much as possible.

### 3. 2. 4. Relation Detection

Our relation set is the coarse-grained relation set proposed by Carlson and Marcu (2001). We chose to stick with the coarse-grained relations, since our corpus is rather small, and also because there would be less errors in the annotation if there were fewer relations. These relations can be found in Table 4.

RST does not allow a span to engage in two rhetorical relations. In these cases, we choose the relation that is more informative or specific (i.e. less frequent) and more in line with the general purpose of the text; e.g. if *Elaboration* and *Cause* hold at the same time, we go with Cause since, in our corpus, it is less frequent and hence more informative or if *Elaboration* and *Cause* hold in an argumentative text, we would prefer *Cause* again since the interpretation in which cause holds is of more argumentative nature compared to *Elaboration* relation. Obviously, to remain as consistent as possible, these preferences must be later on documented and concretely defined.

Figures 5, 6, 7 are some RST trees taken from Persian Rhetorical Structure Theory corpus.

Table 4: Relations in Persian RST Corpus

| Relation | Occurrences (#) | Occurrences (%) |
| --- | --- | --- |
| Span | 3324 | 30.31 |
| Joint | 1927 | 17.57 |
| Elaboration | 1269 | 11.57 |
| Same-Unit | 873 | 7.96 |
| Contrast | 801 | 7.31 |
| Explanation | 396 | 3.61 |
| Attribution | 336 | 3.06 |
| Cause | 527 | 4.81 |
| Background | 249 | 2.27 |
| Evaluation | 316 | 2.88 |
| Topic Comment | 210 | 1.92 |
| Condition | 159 | 1.45 |
| Temporal | 156 | 1.42 |
| Summary | 103 | 0.94 |
| Enablement | 95 | 0.87 |
| Comparison | 98 | 0.89 |
| Topic Change | 82 | 0.75 |
| Manner-Means | 44 | 0.40 |

## 4.     Persian DPLP Parser

Using Persian RST corpus along with Ji and Eisenstein's DPLP parser (2014), we have trained a discourse parser using the Persian RST Treebank. DPLP parser is a parser based on shift-reduce parsing and SVM, and it mainly focuses on projecting lexical features used in discourse parsing.

First, in order to build a discourse segmenter, a binary SVM classifier is trained using the DPLP segmenter module. Each training sample includes the following features for the current token, and tokens located in a window of size 2 around the current token: word, part-of-speech tag, dependency tag; along with the head direction of the current word. Table 5 shows the accuracy, precision, recall and F-measure of our segmenter tested on our ten-document test set. Table 6 presents performance of different discourse segmenters developed so far in comparison with the segmenter trained using Persian RST corpus.

Regarding the parser, DPLP models discourse parsing by combining "large-margin transition-based structured prediction with representation learning" (Ji and Eisenstein, 2014, p. 13), while focusing mainly on modeling surface features by "learning a discourse- driven projection of surface features" by "transforming surface features into a latent space that facilitates RST discourse parsing" (Ji and Eisenstein, 2014, p. 13).

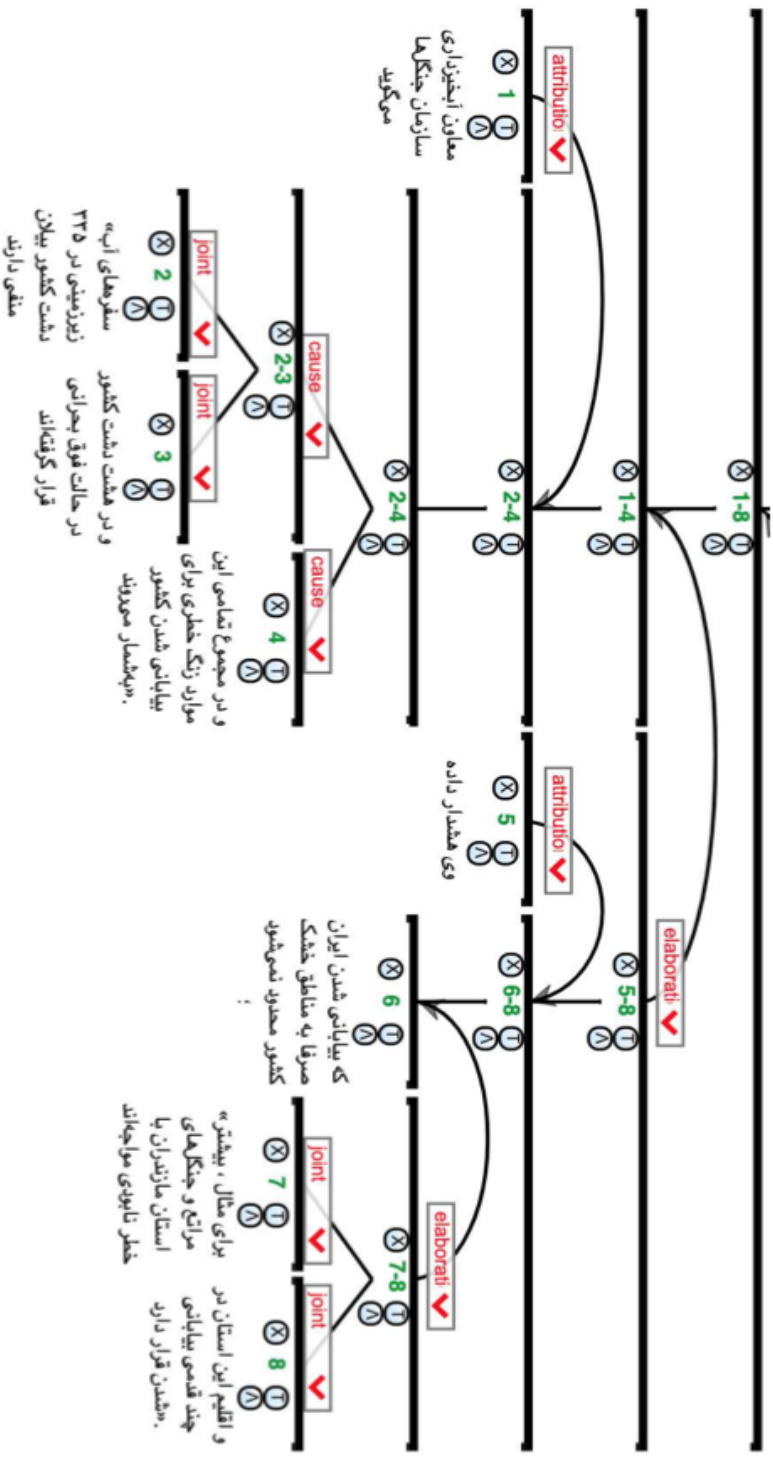

Figure 5: A Partial RST-Tree from Persian Rhetorical Structure Theory Corpus - First Sample

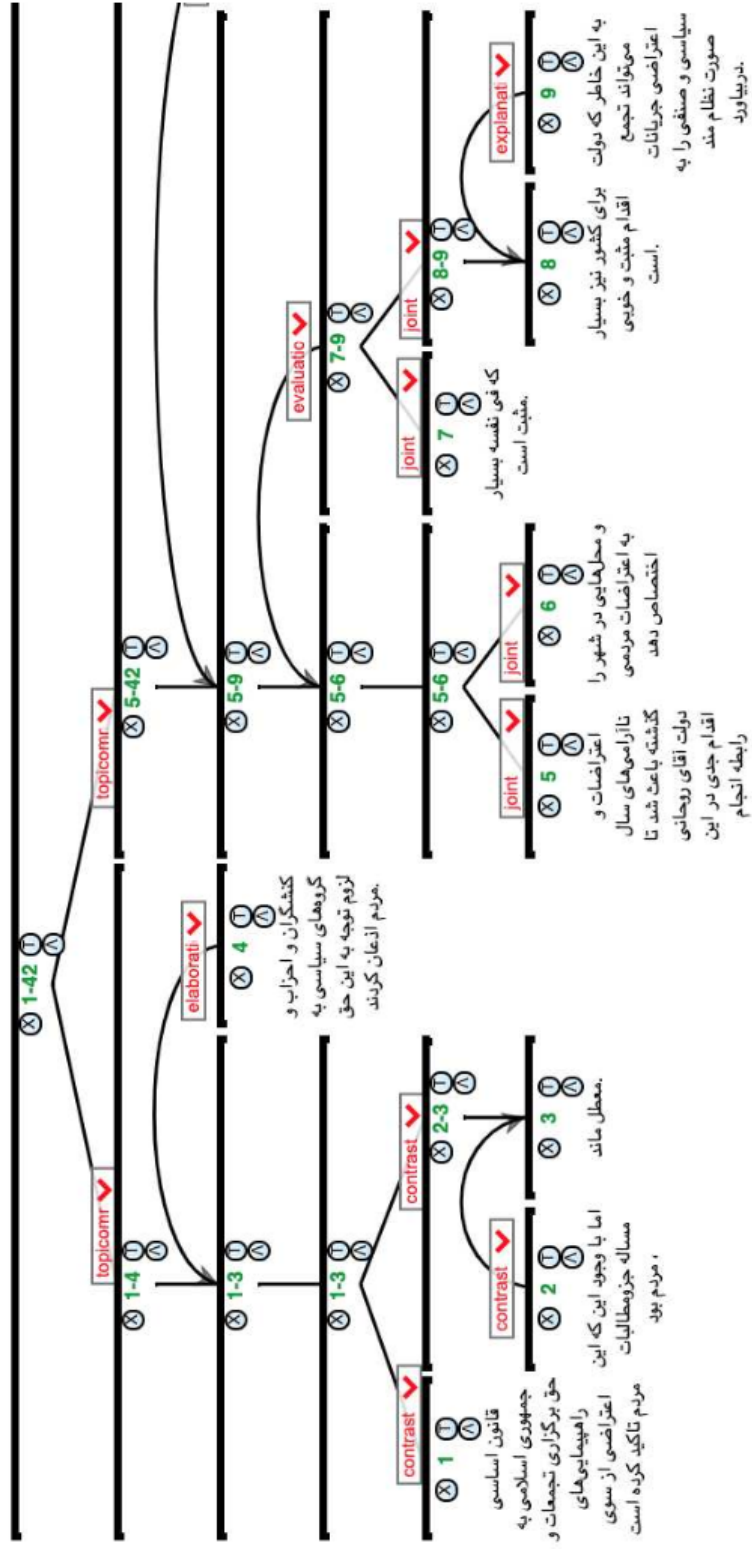

Figure 6: A Partial RST-Tree from Persian Rhetorical Structure Theory Corpus - Second Sample

Figure 7: Another Partial RST-Tree from Persian Rhetorical Structure Theory Corpus - Third Sample

Table 5: Persian Segmenter Performance on Test Set

| Accuracy | Precision | Recall | F1 |
|---|---|---|---|
| 97 | 88 | 85 | 86 |

Table 6: Our Segmenter Performance Compared with English Segmenters

| Segmenter | Precision | Recall |
|---|---|---|
| Sagae (2009) | 87.4 | 86 |
| Hernault et al (2010a) | 95.5 | 94.5 |
| Bourgonje and Schäfer (2019) | 90.2 | 83.5 |
| Bach et al (2012) | 93.1 | 94.2 |
| Hernault et al (2010b) | 96 | 94.6 |
| Wang et al (2018) | 92.9 | 95.7 |
| Our Segmenter | 0.88 | 0.85 |

Learning samples are made by the projected surface feature vectors together with additional feature vectors. Additional features consist of word and its part-of-speech at the beginning of each EDU, length of EDU in token, head wordset from each EDU and so on (Ji and Eisenstein, 2014, p. 18). We have used *stanfordnlp* (Seraji et al., 2012) to get part-of-speech tags, dependency parses and word stems. To get Brown clusters of words, we used the tool provided by Khallash et al (2013).

Table 7: Best Persian DPLP Parsers

| $K$ | $\tau$ | $\lambda$ | S (%) | N (%) | R (%) |
|---|---|---|---|---|---|
| 30 | 0.01 | 1 | 78.70 | 64.01 | 44.28 |
| 30 | 1 | 1 | 78.36 | 63.79 | 44.28 |
| 60 | 0.001 | 50 | 77.80 | 64.13 | 43.72 |
| 30 | 0.001 | 100 | 77.58 | 63.23 | 44.28 |
| 150 | 0.01 | 10 | 77.02 | 62.22 | 42.04 |
| Human* | | | 88.70 | 77.72 | 65.75 |

Following DPLP, to find the best parameters, we performed a grid search -using our development set- over three parameters: λ (SVM regularization parameter) , τ (the pro- jection parameter) and K (the size of projection matrix) by considering the values K ∈ {30, 60, 90, 150}, λ ∈ {1, 10, 50, 100} and τ ∈ {1.0, 0.1, 0.01, 0.001}. Since the training of projection matrix was not included in the DPLP code available at the time, we adapted another implementation[5] of the DPLP algorithm, which adopts the mini-batch learning al- gorithm proposed by Ji and Eisenstein (2014), taking 500 samples of training data in every iteration. We learned the concatenation form of the projection matrix and used gold EDU segmentation to train the parser. Table 7 presents the performance of our top-five models. Table 8 is a comparison of different RST parser built so far, including ours. Table 9 shows the confusion matrix of relation detection in Persian DPLP parser for the ten most common relations.

Table 8: A Comparison of Different RST Parsers

| Parser | S (%) | N (%) | R (%) |
|---|---|---|---|
| Persian DPLP (This Work) | 78.70 | 64.01 | 44.28 |
| Hernault et al (2010) | 83 | 68.4 | 54.8 |
| Joty et al (2013) | 82.47 | 68.43 | 57.63 |
| Ji & Eisenstein (2014) | 82.08 | 71.13 | 61.63 |
| Li et al (2014) | 84 | 70.8 | 58.6 |
| Feng & Hirst (2014a) | 83 | 68.4 | 54.8 |
| Heilman & Sagae (2015) | 82.6 | 67.1 | 55.4 |
| Liu et Lapata (2017) | 85.8 | 71.1 | 58.9 |
| Wang et al (2017) | 86 | 72.4 | 59.7 |
| Human* | 88.70 | 77.72 | 65.75 |

5. **Conclusion and Future Works**

Through this paper, we have presented Persian RST corpus as well as a Persian discourse parser. This corpus consists of a total of 150 texts. Annotation has been done by mainly following the English *RST Discourse Treebank*. Our discourse parser was trained and built using the DPLP algorithm. Although the performance of this parser is far from perfect, it looks like building better discourse parsers for Persian would be a feasible project in the future. Better RST parsers can be made by increasing the training data, adding more texts to the corpus, and reducing inconsistencies in annotation as well as trying out other approaches and algorithms to implement RST parsers.

---

[5] https://github.com/levyrafi12/RST

Table 9: Relation Detection Confusion Matrix

|  | Span | Joint | Elaboratiob | Same-Unit | Contrast-NN | Explanation | Attribution | Cause-NN | Cause | Background |
|---|---|---|---|---|---|---|---|---|---|---|
| Span | 60.77 | 20.10 | 6.22 | 6.70 | 2.39 | 0.48 | 1.91 | 0.00 | 0.00 | 1.44 |
| Joint | 1.39 | 73.38 | 3.90 | 2.60 | 6.49 | 2.60 | 0.00 | 0.00 | 0.00 | 0.65 |
| Elaboratiob | 7.79 | 15.58 | 75.32 | 1.30 | 0.00 | 0.00 | 0.00 | 0.00 | 0.00 | 0.00 |
| Same-Unit | 7.02 | 1.75 | 0.00 | 87.72 | 0.00 | 0.00 | 3.51 | 0.00 | 0.00 | 0.00 |
| Contrast-NN | 18.18 | 24.24 | 6.06 | 3.03 | 45.45 | 0.00 | 0.00 | 0.00 | 0.00 | 3.03 |
| Explanation | 40.00 | 33.33 | 20.00 | 0.00 | 0.00 | 6.67 | 0.00 | 0.00 | 0.00 | 3.70 |
| Attribution | 22.22 | 3.70 | 0.00 | 3.70 | 0.00 | 0.00 | 66.67 | 0.00 | 0.00 | 0.00 |
| Cause-NN | 15.38 | 61.54 | 15.38 | 0.00 | 0.00 | 0.00 | 7.69 | 0.00 | 0.00 | 0.00 |
| Cause | 25.00 | 60.00 | 5.00 | 0.00 | 0.00 | 5.00 | 5.00 | 0.00 | 0.00 | 0.00 |
| Background | 35.71 | 21.43 | 14.29 | 7.14 | 0.00 | 0.00 | 0.00 | 0.00 | 7.14 | 14.29 |

# References

Paula CF Cardoso, Erick G Maziero, Mara Luca Castro Jorge, Eloize MR Seno, Ariani Di Felippo, Lucia Helena Machado Rino, Maria das Gracas Volpe Nunes, and Thiago AS Pardo. Cstnews-a discourse-annotated corpus for single and multi-document summa- rization of news texts in brazilian portuguese. In Proceedings of the 3rd RST Brazilian Meeting, pages 88–105, 2011. URL http://www.nilc.icmc.usp.br/nilc/download/ ariani/CardosoETAL_RST_2011.pdf.

Lynn Carlson and Daniel Marcu. Discourse tagging reference manual. ISI Technical Re- port ISI-TR-545, 54:56, 2001. URL http://faculty.washington.edu/fxia/lsa2011/ readings/RST-tagging-ref-manual.pdf.

Lynn Carlson, Daniel Marcu, and Mary Ellen Okurowski. Building a discourse-tagged corpus in the framework of rhetorical structure theory. In Current and new directions in discourse and dialogue, pages 85–112. Springer, 2003. URL https://link.springer. com/chapter/10.1007/978-94-010-0019-2_5.

Iria Da Cunha, Juan-Manuel Torres-Moreno, and Gerardo Sierra. On the development of the rst spanish treebank. In Proceedings of the 5th Linguistic Annotation Workshop, pages 1–10, 2011. URL https://www.aclweb.org/anthology/W11-0401.

Debopam Das and Manfred Stede. Developing the bangla rst discourse treebank. In Pro- ceedings of the Eleventh International Conference on Language Resources and Evaluation (LREC 2018), 2018. URL http://www.lrec-conf.org/proceedings/lrec2018/pdf/ 387.

Vanessa Wei Feng and Graeme Hirst. A linear-time bottom-up discourse parser with con- straints and post-editing. In Proceedings of the 52nd Annual Meeting of the Association for Computational Linguistics (Volume 1: Long Papers), pages 511–521, 2014a. URL https://www.aclweb.org/anthology/P14-1048.

Vanessa Wei Feng and Graeme Hirst. Two-pass discourse segmentation with pairing and global features. arXiv preprint arXiv:1407.8215, 2014b. URL https://arxiv.org/abs/ 1407.8215.

Michael Heilman and Kenji Sagae. Fast rhetorical structure theory discourse parsing. arXiv preprint arXiv:1505.02425, 2015. URL https://arxiv.org/abs/1505.02425.

Hugo Hernault, Helmut Prendinger, Mitsuru Ishizuka, et al. Hilda: A discourse parser using support vector machine classification. Dialogue & Discourse, 1(3), 2010. URL http://hernault.fr/pubs/dd2010-hugo.

Mikel Iruskieta, Marıa J Aranzabe, Arantza Diaz de Ilarraza, Itziar Gonzalez, Mikel Ler- sundi, and Oier Lopez de Lacalle. The rst basque treebank: an online search inter- face to check rhetorical relations. In 4th workshop RST and discourse studies, pages 40–49, 2013. URL https://ixa.ehu.eus/sites/default/files/dokumentuak/3960/ 2013RST-Basque-TB.pdf.

Yangfeng Ji and Jacob Eisenstein. Representation learning for text-level discourse pars- ing. In Proceedings of the 52nd annual meeting of the association for computational linguistics (volume 1: Long papers), pages 13–24, 2014. URL https://www.aclweb.org/

anthology/P14-1002.

Marianne W Jorgensen and Louise J Phillips. Discourse analysis as theory and method. Sage, 2002.

Shafiq Joty, Giuseppe Carenini, Raymond Ng, and Yashar Mehdad. Combining intra-and multi-sentential rhetorical parsing for document-level discourse analysis. In Proceedings of the 51st Annual Meeting of the Association for Computational Linguistics (Volume 1: Long Papers), pages 486–496, 2013. URL https://www.aclweb.org/anthology/ P13-1048.

Mojtaba Khallash, Ali Hadian Cefidekhanie, and Behrouz Minaei-Bidgoli. An empirical study on the effect of morphological and lexical features in persian dependency parsing. In Proceedings of the Fourth Workshop on Statistical Parsing of Morphologically-Rich Languages, pages 97–107, 2013. URL https://www.aclweb.org/anthology/W13-4912.

Jiwei Li, Rumeng Li, and Eduard Hovy. Recursive deep models for discourse parsing. In

Proceedings of the 2014 Conference on Empirical Methods in Natural Language Processing (EMNLP), pages 2061–2069, 2014. URL https://www.aclweb.org/anthology/ D14-1220.

Yang Liu and Mirella Lapata. Learning contextually informed representations for linear-time discourse parsing. In Proceedings of the 2017 Conference on Empirical Methods in Natural Language Processing, pages 1289–1298, 2017. URL https://www.aclweb.org/anthology/D17-1133/.

William C Mann and Sandra A Thompson. Rhetorical structure theory: A theory of text or- ganization. University of Southern California, Information Sciences Institute Los Angeles, 1987. URLhttp://www.sfu.ca/rst/05bibliographies/bibs/ISI_RS_87_190.pdf.

William C Mann and Sandra A Thompson. Rhetorical structure theory: Toward a functional theory of text organization. Text, 8(3):243–281, 1988. URL https://www.cl.cam.ac.uk/teaching/1617/R216/rst.pdf.

YUE Ming. Rhetorical structure annotation of chinese news commentaries. Journal of Chinese Information Processing, 4, 2008.

Azadeh Mirzaei and Pegah Safari. Persian discourse treebank and coreference corpus. In

Proceedings of the Eleventh International Conference on Language Resources and Evalu- ation (LREC 2018), 2018. URL https://www.aclweb.org/anthology/L18-1638.

Livia Polanyi, Chris Culy, Martin Van Den Berg, Gian Lorenzo Thione, and David Ahn. A rule based approach to discourse parsing. In Proceedings of the 5th SIGdial Workshop on Discourse and Dialogue at HLT-NAACL 2004, pages 108–117, 2004. URL https://www.aclweb.org/anthology/W04-2322.

Mohammad Sadegh Rasooli, Amirsaeid Moloodi, Manouchehr Kouhestani, and Behrouz Minaei-Bidgoli. A syntactic valency lexicon for persian verbs: The first steps towards persian dependency treebank. In 5th Language & Technology Conference (LTC): Human Language

Technologies as a Challenge for Computer Science and Linguistics, pages 227– 231, 2011. URL https://www.researchgate.net/publication/230612993.

Mojgan Seraji, Beáta Megyesi, and Joakim Nivre. Dependency parsers for persian. In 24th International Conference on Computational Linguistics, 8-15 December, 2012, Mumbai, India. ACL Anthology, 2012. URL https://www.aclweb.org/anthology/W12-5205.

Manfred Stede. The potsdam commentary corpus. In Proceedings of the Workshop on Discourse Annotation, pages 96–102, 2004. URL https://www.aclweb.org/anthology/W04-0213.

Manfred Stede. Discourse processing. Synthesis Lectures on Human Language Technologies, 4(3):1–165, 2011. URL https://www.morganclaypool.com/doi/abs/10.2200/s00354ed1v01y201111hlt015.

Omid Tabibzadeh. Verb Valency and Sentence Base Structures in Modern Persian. Markaz, 2006.

Maite Taboada and William C Mann. Applications of rhetorical structure theory. Discourse studies, 8(4):567–588, 2006. URL https://doi.org/10.1177/1461445606064836.

Svetlana Toldova, Dina Pisarevskaya, Margarita Ananyeva, Maria Kobozeva, Alexander Nasedkin, Sofia Nikiforova, Irina Pavlova, and Alexey Shelepov. Rhetorical relations markers in russian rst treebank. In Proceedings of the 6th Workshop on Recent Advances in RST and Related Formalisms, pages 29–33, 2017. URL https://www.aclweb.org/anthology/W17-3604/.

Nynke Van Der Vliet, Ildikó Berzlánovich, Gosse Bouma, Markus Egg, and Gisela Redeker. Building a discourse-annotated dutch text corpus. Bochumer Linguistische Arbeitsberichte, 3:157–171, 2011. URL https://d-nb.info/1119241413/34.

Yizhong Wang, Sujian Li, and Houfeng Wang. A two-stage parsing method for text-level discourse analysis. In Proceedings of the 55th Annual Meeting of the Association for Computational Linguistics (Volume 2: Short Papers), pages 184–188, 2017. URL https://www.aclweb.org/anthology/P17-2029.

Siamak Zandrazavi. Opponents of identity, May 2019. URL https://www.magiran.com/article/3901734.

Amir Zeldes. rstweb-a browser-based annotation interface for rhetorical structure theory and discourse relations. In Proceedings of the 2016 Conference of the North American Chapter of the Association for Computational Linguistics: Demonstrations, pages 1–5, 2016. URLhttps://www.aclweb.org/anthology/N16-3001/.

Amir Zeldes. The gum corpus: Creating multilayer resources in the classroom. Language Resources and Evaluation, 51(3):581–612, 2017. URL https://link.springer.com/article/10.1007/s10579-016-9343-x.